\documentclass{article}
\usepackage{ijcai22}

\usepackage{times}
\usepackage{soul}
\usepackage{url}
\usepackage[utf8]{inputenc}
\usepackage[small]{caption}
\usepackage{graphicx}
\usepackage{amsmath}
\usepackage{amsthm}
\usepackage{booktabs}
\usepackage{amssymb}
\usepackage{amsmath}
\usepackage{url}
\usepackage{array}
\usepackage{booktabs}
\usepackage[colorlinks]{hyperref}
\usepackage{verbatim}
\usepackage{flushend}

\usepackage{algorithmicx}
\usepackage{algorithm}
\usepackage[noend]{algpseudocode}

\algnewcommand\algorithmicinput{\textbf{Assume:}}
\algnewcommand\Assume{\item[\algorithmicinput]}

\title{A multi-task semi-supervised framework for Text2Graph \& Graph2Text}

\author{
Oriol Domingo$^1$\and
Marta R. Costa-Jussà$^2$\and
Carlos Escolano$^{2}$
\affiliations
$^1$BATOU XYZ\\
$^2$Universitat Politècnica de Catalunya\\
\emails
oriol@batou.xyz,
\{marta.ruiz, carlos.escolano\}@upc.edu}

\begin{document}

\maketitle

\begin{abstract}
The Artificial Intelligence industry regularly develops applications that mostly rely on Knowledge Bases, a data repository about specific, or general, domains, usually represented in a graph shape. Similar to other databases, they face two main challenges: information ingestion and information retrieval. We approach these challenges by jointly learning graph extraction from text and text generation from graphs. The proposed solution, a T5 architecture, is trained in a multi-task semi-supervised environment, with our collected non-parallel data, following a cycle training regime. Experiments on WebNLG dataset show that our approach surpasses unsupervised state-of-the-art results in text-to-graph and graph-to-text. More relevantly, our framework is more consistent across seen and unseen domains than supervised models. The resulting model can be easily trained in any new domain with non-parallel data, by simply adding text and graphs about it, in our cycle framework.\footnote{Source code is available under request and non-paralled data is released at: \url{https://github.com/uridr/GTWiki}}
\end{abstract}

\section{Introduction}

In recent years, Artificial Intelligence industry has leveraged the power of computation along deep learning models to build cutting-edge applications. Some of these applications, such as question answering systems \cite{SaiSharath2021ConversationalQA}, chat-bot \cite{lili2020}, recommender systems \cite{9216015} or personal assistants \cite{bellegarda2013large}, heavily rely on Knowledge Bases (KB). A KB is a database that stores complex structured and unstructured facts about specific, or generic, areas, commonly expressed in a graph shape. The following challenges arise in the aforementioned applications when dealing with KBs:

\begin{figure}[t]
    \centering
    \includegraphics[scale=0.3]{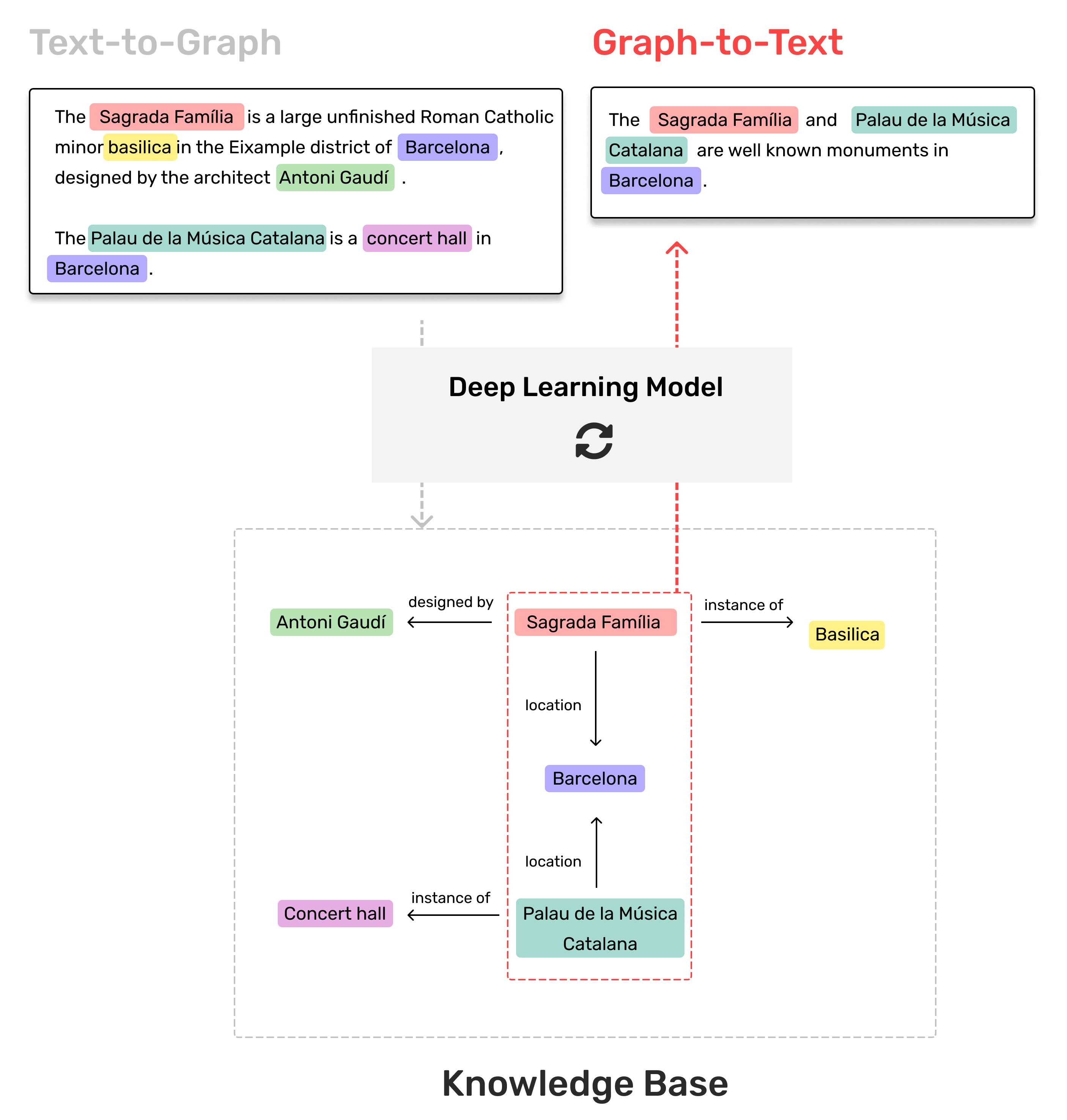}
    \caption{Overview of the problem. A Deep Learning model jointly learns Text-to-Graph (gray) and Graph-to-Text (red) in a cycle training regime.}
    \label{fig:overview}
\end{figure}

\begin{enumerate}
    \item Continuous information ingestion to keep the KB updated according to the (world) situation \cite{ji2011knowledge}.
    \item Building a data representation on top of the information retrieval level that is human comprehensible \cite{ferreira20202020}.
\end{enumerate}

To address these challenges, we apply Text-to-Graph (T2G) to uncover new hidden facts given a textual source (1st challenge). For instance, the information embedded in the text \textit{"The Palau de la Música Catalana is a concert hall in Barcelona"} can be summarised as  \textit{“(Palau de la Música Catalana, instance of, Concert Hall), (Palau de la Música Catalana, location, Barcelona)”}. Besides, we perform Graph-to-Text (G2T) to generate text that embeds the whole information of the retrieved knowledge graph (2nd challenge). For example, the graph \textit{“(Sagrada Família, location, Barcelona), (Palau de la Música Catalana, location, Barcelona)”} can be expressed as \textit{“The Sagrada Família and  Palau de la Música Catalana are well known monuments located in Barcelona.”} By approaching both tasks together, we can consider them as complementary tasks [Figure \ref{fig:overview}].

Despite being very well known tasks, there are still very little datasets, and those are on the magnitude of tens of thousands samples \cite{ferreira20202020}, making it hard for models to reach human-level performance. For this reason, our work investigates a framework that approaches both tasks in a multi-task semi-supervised environment, aiming to gain data efficiency, reduce overfitting through shared representations, and fast learning by leveraging auxiliary information  \cite{crawshaw2020multi}.

%Recent works \cite{agarwal2020machine} \cite{dognin2021regen} solve both, T2G and G2T, in a multi-task set-up, aiming to gain data efficiency, reduce overfitting through shared representations, and fast learning by leveraging auxiliary information \cite{crawshaw2020multi}. Simultaneously, \cite{guo2020cyclegt} propose a cycle training environment for unsupervised learning, through learning from a a large number of unlabeled samples. This work investigates both techniques in combination, i.e. solving T2G and G2T in a multi-task semi-supervised environment.

To summarise, our main contributions are two fold:

\begin{itemize}
    \item Building a model capable of learning G2T and T2G in a multi-task semi-supervised environment. Experiments on WebNLG dataset show that our framework surpasses state-of-the-art results in G2T \& T2G for unsupervised models. More relevantly, our framework is on par or even more consistent across seen and unseen domains than supervised and unsupervised models on both tasks.
    \item Collecting and releasing non-parallel data, text and graphs, for unsupervised models of around 250k instances per each data type.
\end{itemize}

The rest of the paper is organised as follows. In section \ref{problem-formulation} both challenges are mathematically formulated, before related work is presented in section \ref{related-work}. Following, the methodology of this work is explained in section \ref{methodology}, where we detail each part of our framework. After that, section \ref{experiments} introduces our experimental set-up, and then, our results on these experiments are discussed in section \ref{results}. Finally, conclusions are presented in section \ref{conclusions}.

\section{Problem Formulation}\label{problem-formulation}

We formulate information ingestion and information retrieval as a T2G (parsing) and G2T (generation) tasks respectively, using a dataset of supervised examples $\mathcal{S}: (g,t)_{i=1}^N\in\mathcal{G}\times\mathcal{T}$. On one hand, our graph dataset $\mathcal{G}:=\{g_i\}_{i=1}^N$ consists of $N$ graphs in a triple format, where triples are composed of a subject ($s_n$), predicate ($p_n$) and object ($o_n$). On the other hand, our text corpus $\mathcal{T}:=\{t_i\}_{i=1}^N$ consists of $N$ text sequences, made up by tokens. Presumably, aligned graphs and sequences on $\mathcal{S}$ share the same latent content, but differ in terms of surface realisation. Thus, we aim to generate text from graphs using some function $g2t_\theta$, and inversely, we aim to parse graphs from text using some function $t2g_\phi$, where each output embeds the semantic meaning of the aligned input [Eq \ref{meth-eq-3}]. Both functions are parameterized by $\theta$ and $\phi$ respectively.
% We should mention a table with textual data instead of this

\begin{equation}\label{meth-eq-3}
    \begin{cases} 
          g2t_\theta(g) = \hat{t} \simeq t \\
          t2g_\phi(t) =  \hat{g} \simeq g
     \end{cases}
\end{equation}\\

Ideally, these functions are optimised over $\mathcal{S}$ by means of a maximum log-likelihood estimation on $\theta$ and $\phi$[Eq \ref{meth-eq-4}]. By this optimisation pass, models aim to learn the corresponding T2G and G2T tasks.

\begin{equation}\label{meth-eq-4}
     \mathcal{J}(\theta,\phi)=\mathbb{E}_{(g,t)\sim \mathcal{S}}[\log \ p(t\shortmid g;\theta)  +  \log \ p(g\shortmid t;\phi) ]
\end{equation}

\section{Related Work}\label{related-work}

\setlength{\parindent}{0pt}\textbf{Unsupervised Learning}. Recently, \cite{schmitt2020unsupervised} presented the first approach to unsupervised G2T and T2G. They proposed two different methods: a rule-based system; they considered as a baseline; and a neural sequence-to-sequence system; which considerably improved baseline results. On one hand, the rule-based system relies on several steps such as preprocessing, removing stop words, part of speech tagging (similar to semantic parsers), heuristics and template linearisation. On the other hand, they proposed a BiLSTM \cite{hochreiter1997long} sequence-to-sequence system trained also in a multi-task environment and fine-tuned with noisy source samples \cite{veit2017learning}. Regarding the neural training regime, firstly, they obtained a language model for both graphs and text, and later on iterative Back Translation \cite{sennrich2015improving} is applied on graphs and text in which their corresponding alignment has been removed, but come from a supervised dataset in which the system is evaluated on. Similarly, \cite{guo2020cyclegt} developed an unsupervised training method that can bootstrap from fully non-parallel graphs and text data, and iteratively back translate between the two forms. However, unsupervised samples come from a supervised dataset as well, so a real data distribution exists among their training examples unlike in our work. This cycle training framework achieves state-of-the-art results for unsupervised models in this domain, however, they rely on two different models. Particularly, T2G task is solved with a system made up by an off-the-shelf entity extraction model \cite{qi2020stanza} and a BiLSTM to predict the relation between any pair of entities. Shortly, G2T is solved with a T5 model \cite{raffel2019exploring}, as we do (see \ref{multi-task}). Consequently, not only does this approach seek to have a common data representation between these two models, but this also constraints the optimisation procedure as both models cannot be updated together, rather they are updated one-by-one, making loss function non-differentiable.

\setlength{\parindent}{0pt}\textbf{Multi-task Learning}\label{rw-multitask}. As previously mentioned, Martin Schmitt et al. trained their neural model in a multi-task set-up sharing encoder and decoder. However, they tell the decoder which type of output should be produced (text or graphs) by means of cell state initialisation in the decoder side, with an embedding corresponding to the desired output type. Alternatively, Oshin Agarwal et al. \cite{agarwal2020machine} overcome the need of task specification at decoder level thanks to the input format of the chosen model. The model is also a pre-trained T5 model, in which the task to be solved is simply identified by adding task tokens to the input. Moreover, they build a bilingual model for English and Russian languages. Stated by the authors, these capabilities remarkably improves on unseen relations and Russian. Apart from that, they also perform data augmentation but not in an unsupervised manner, they rather pre-trained the original model on a parallel corpus of news data: WMT-News corpus \cite{tiedemann2012parallel}. %\cite{dognin2021regen} present a hybrid model based on T5-Large architecture. More relevantly, they show that reinforcement learning along self-critical sequence training benefits graph and text generation, leading to reach state-of-the-art on both tasks for supervised models.

\begin{figure*}[t]
    \centering
    \includegraphics[scale=0.45]{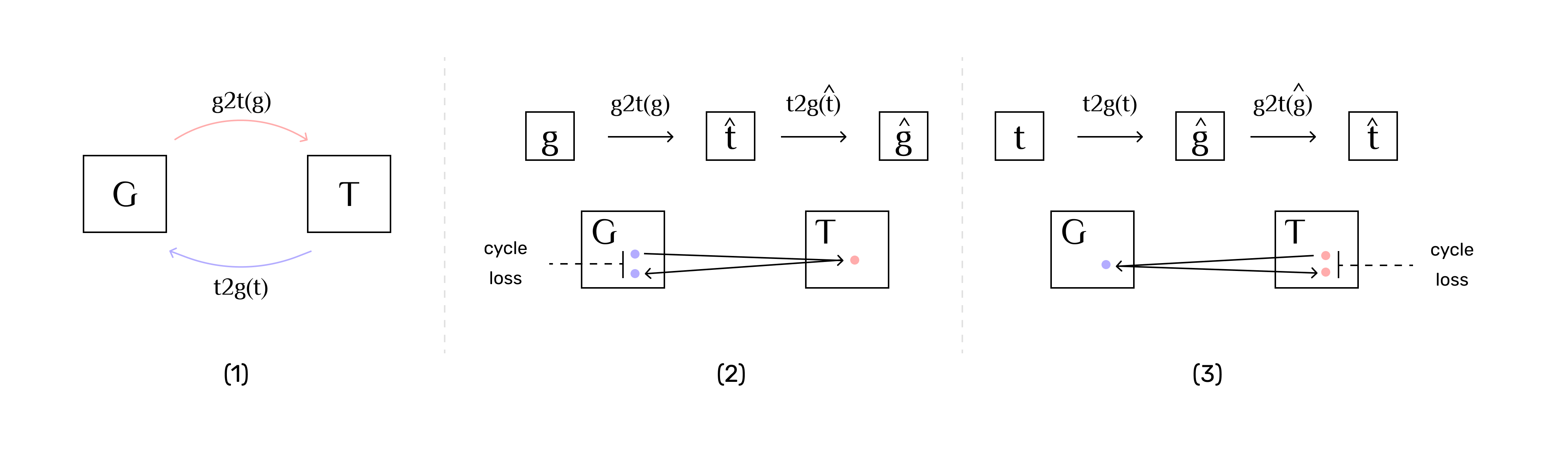}
    \caption{(1) The cycle training approach is based on two mapping functions  $g2t:\mathcal{G} \rightarrow \mathcal{T}$ and $t2g : \mathcal{T} \rightarrow \mathcal{G}$, however, our approach uses a single model to learn both functions. This framework introduces two cycle losses that capture the intuition that if we translate from one side to the other, and back again we should obtain the original sample: (2) forward cycle loss: $g \rightarrow g2t(g) \rightarrow t2g(g2t(g)) \simeq g$, and (3) backward cycle loss: $t \rightarrow t2g(t) \rightarrow g2t(t2g(t)) \simeq t$.}
    \label{fig:cycle}
\end{figure*}

\section{Methodology}\label{methodology}
In this section, we will present our approach towards multi-task learning and unsupervised learning. Afterwards, the optimisation procedure is detailed.

\subsection{Multi-task Learning}\label{multi-task}
We propose a multi-task environment, where a single model ($z_\alpha$), parameterized by $\alpha$, can learn both functions ($g2t_\theta$ and $t2g_\phi$) simultaneously, leading to [Eq \ref{meth-eq-5}]. Thereafter, our main challenge is to approximate [Eq \ref{meth-eq-5}] with unsupervised samples using the model itself through a cycle training framework (see \ref{unsupervised}).

\begin{equation}\label{meth-eq-5}
    \mathcal{J}(\alpha)=\mathbb{E}_{(g,t)\sim \mathcal{S}}[\log \ p(t \shortmid g;\alpha) \ + \log \ p(g \shortmid t;\alpha)]
\end{equation}

We are going to use the pre-trained T5-Base model \cite{raffel2019exploring}. This model is a Transformer architecture, that it is roughly equivalent to the original one \cite{vaswani2017attention}, however, it introduces an unified framework that converts all text-based language problems into text-to-text format by means of a task token specification. This framework provides a consistent training objective for both, pre-training and fine-tuning, being the latter of special interest in our work as it eases the optimisation of our single model ($z_\alpha$) on both downstream tasks at the same time. 

\subsection{Unsupervised Learning}\label{unsupervised}Cycle training was originally suggested as an image-to-image translation, rather than text-to-text (our current approach), a problem where the goal is to learn a mapping between an input image and an output image \cite{zhu2017unpaired}. It was presented as a solution to the absence of paired examples. This solution was based on a cycle consistency loss relying on Generative Adversarial Networks \cite{goodfellow2014generative}.

The main constraint for using cycle training is that there must exist two complementary tasks that guarantees that the input of one task is the output of the other task, and vice-versa [Figure \ref{fig:cycle} (1)]. For instance, a graph can be fed into a model to generate some text. The resulting text  can  also be fed into this model to generate a graph, which should resemble the original one [Figure \ref{fig:cycle} (2)]. The same procedure is applied in the reverse direction, i.e. starting from text and generating synthetic graph [Figure \ref{fig:cycle} (3)]. At this point, we can cycle-train our model since a reference of our hypothesis exists on both sides. These steps constitute the iterative loop in which cycle training is based on.

In our case, previous constraint holds, existence of complementary tasks, so it is possible to build a bijective mapping function that given a variable $x$ satisfies $x = t2g(g2t(x))$, where $g2t$ is the inverse function of $t2g$. However, we approach both tasks using a single model ($z_\alpha$), following a multi-task set-up,  hence, it must hold that $z:= g2t = t2g^{-1} = t2g$, which is an involutory function. This constraint needs to be slightly relaxed as our variable $x$, representing text or graphs, is concatenated with an extra token ($w$ and $k$)\footnote{These are task tokens for our T5 model.} at the beginning of the sentence in order to specify the model which output should be generated, and the output does not contain this extra token to again specify the model which task is solving. Thus, the input vector space is marginally modified with respect to the original input, text or graph, before passing into the model [Eq \ref{meth-eq-6}]. 

\begin{equation}\label{meth-eq-6}
    x \simeq \hat{x} =z(z(x)) \longrightarrow  x \simeq \hat{x} = z(w \ \parallel \ (z(k \ \parallel \ x)) 
\end{equation}

This space modification helps the model to avoid learning the identity function - the most simple involutory function - because the input of the function is slightly modified with respect to the output of the function due to task token specification. Furthermore, we use different data on each iteration of the cycle training (see \ref{exp:training}) to prevent our model to memorize all the training data regardless being synthetic or real, and not generalize well on test data.

\begin{table*}[!t]
\centering 
\scalebox{0.8}{
    \begin{tabular}{l c c c  c c c c c c c c c}
    \toprule
      {} & \multicolumn{3}{c}{\textbf{Overall}} &  \multicolumn{3}{c}{\textbf{Seen Categories}} & \multicolumn{3}{c}{\textbf{Unseen Entities}} & \multicolumn{3}{c}{\textbf{Unseen Categories}} \\
      \\
      {}  & BLEU & TER  & chrF++ & BLEU & TER  & chrF++ & BLEU & TER  & chrF++ & BLEU & TER  & chrF++ \\
    \midrule
    %ReGen ($^{\diamondsuit \triangle}$) \cite{dognin2021regen}  &  56.3  &  -   & 0.70 & - & - & - & - & - & - & - & - & - \\
    bt5 ($^{\diamondsuit \triangle}$) \cite{agarwal2020machine}   &  51.7 & 0.43 & 0.67 & 61.1 & 0.39 & 0.72 & 50.8 & 0.41 & 0.68 & 44.0 & 0.47 & 0.63 \\
    CycleGT ($^{\heartsuit}$) \cite{guo2020cyclegt}   &   44.6 & 0.47 & 0.63  & 47.4 & 0.49 & 0.65 & 46.6 & 0.44 & 0.65 & 40.9 & 0.48 & 0.61 \\
    \midrule
    T5-Baseline ($^{\diamondsuit \triangle}$) [our implementation] & 44.6 & 0.51 & 0.54 & 51.7 & 0.49 & 0.57 & 45.4 & 0.49 & 0.55 & 38.0 & 0.52 & 0.51 \\
    T5-Finetune ($^{\diamondsuit \triangle}$) [our implementation] & 42.5 & 0.51 & 0.60 &  46.5 & 0.52 & 0.61 & 44.4 & 0.48 & 0.63 & 38.2 & 0.51 & 0.57 \\
    \textbf{T5-MSSF} ($^{\diamondsuit \Join}$) [our framework] & 45.3 & 0.48 & 0.62 & 50.9 & 0.49 & 0.64 & 45.8 & 0.45 & 0.64 & 40.0 & 0.49 & 0.59 \\
    \bottomrule
    \end{tabular}
}
\caption{Multi-task (${\diamondsuit}$). Unspervised (${\heartsuit}$). Semi-Supervised ($\Join$). Supervised (${\triangle}$). Summary of related work results in Graph-to-Text (G2T) with multi-task, supervised and/or unsupervised learning. Our multi-task semi-supervised framework is T5-MSSF.}
\label{tab:g2t-results}
\end{table*}

\begin{table}[!t]
    \centering
    \begin{tabular}{l c c}
        \toprule
        & \textbf{Test} (G2T) & \textbf{Test} (T2G)   \\
        \midrule
         Seen categories    & 490 (28\%) &  606 (28\%)    \\
         Unseen entities    & 393 (22\%) &  457 (21\%)    \\
         Unseen categories  & 896 (50\%) &  1,092 (51\%)  \\
         \midrule
         Total              & 1,779      &  2,155         \\
        \bottomrule
    \end{tabular}
    \caption{Number of the test instances for Graph-to-Text (G2T) and Text-to-Graph (T2G) with respect to the different data types.}
    \label{tab:test-split-stats}
\end{table}

\subsection{Optimisation Procedure}

This mathematical framework allows training without or with few parallel data. The main idea is that the model can learn from unlabeled data: unlabeled graphs  $\mathcal{U}_\mathcal{G}$ and unlabeled text $\mathcal{U}_{\mathcal{T}}$; using its own predictions [Eq \ref{meth-eq-7} left] as reference during training [Eq \ref{meth-eq-7} right]. The optimisation pass for both cycle losses can be backpropagated together on each batch [Eq \ref{meth-eq-8}].

\begin{align}\label{meth-eq-7}
    \begin{cases} 
          z_\alpha(t)=\hat g \longrightarrow \mathcal{L}_{cycleG2T}=\mathbb{E}_{t\in\mathcal{U}_{\mathcal{T}}}[ \ -\log \ p(t\shortmid\hat g;\alpha) \ ] \\
          z_\alpha(g)=\hat t \longrightarrow \mathcal{L}_{cycleT2G}=\mathbb{E}_{g\in\mathcal{U}_{\mathcal{G}}}[ \ -\log \ p(g\shortmid\hat t;\alpha) \ ] 
     \end{cases}
\end{align}

\begin{align}\label{meth-eq-8}
    \mathcal{L}_{cycle}=\mathcal{L}_{cycleG2T} + \mathcal{L}_{cycleT2G}
\end{align}

The [Eq \ref{meth-eq-8}] resembles the supervised one [Eq \ref{meth-eq-4}], however, with predicted inputs from the unlabeled samples. Notice here that there is no-alignment between graphs and text from both $\mathcal{U}_\mathcal{G}$ and $\mathcal{U}_{\mathcal{T}}$. On every cycle step, our model uses its predictions as inputs, which are synthetic and so can change on every prediction. This variability helps the model to avoid over-fitting training data.

Before training our model ($z_\alpha$) on the unsupervised data, we have to apply a fine-tuning step to learn a representation of these tasks. To do so, we optimise our model, over $\alpha$, on a maximum likelihood estimation [Eq \ref{meth-eq-9}], in which both tasks are simultaneously considered - following our multi-task approach.

\begin{align}\label{meth-eq-9}
    \alpha^* = argmax_\alpha \prod_{(g,t)\sim S} p(t\shortmid g;\alpha) \cdot p(g\shortmid t;\alpha)
\end{align}

Finally, we can train our model on the unsupervised samples through the cycle training framework. On every cycle training step, the parameters of the model are optimised using a maximum likelihood estimation [Eq \ref{meth-eq-10}], where synthetic samples are the inputs, but human written text and graphs are the reference respectively.

\begin{align}\label{meth-eq-10}
    \alpha^* = argmax_\alpha \prod_{(\hat g,t)\sim \mathcal{U}_{\mathcal{T}}} p(t\shortmid \hat g;\alpha) \prod_{(g,\hat t)\sim \mathcal{U}_{\mathcal{G}}} p(g\shortmid \hat t;\alpha)
\end{align}

\section{Experimental Framework}\label{experiments}

\setlength{\parindent}{0pt}\textbf{Dataset}. The WebNLG corpus \cite{gardent2017creating} is a common benchmark on which to evaluate and compare G2T \& T2G systems\footnote{\url{https://gitlab.com/shimorina/webnlg-dataset}}. In 2020, they released the latest version of the corpus \cite{ferreira20202020}, which included Russian language and data for T2G (parsing). The latest English version of the corpus is the 3.01 version, in which the training and development sets comprise 16 different DBpedia categories. Apart from that, the test sets have three different data types:

\begin{itemize}
    \item \textbf{Seen categories}: triples containing the entities and categories seen in the training data.
    \item \textbf{Unseen entities}: triples containing the categories seen in the training data, but not entities.
    \item \textbf{Unseen categories}: triples containing the categories not present in the training data.
\end{itemize}

Test split's statistics for both tasks are shown in [Table \ref{tab:test-split-stats}].%and few examples are showcased in [Table \ref{tab:sr-predicts} \ref{tab:re-predicts}]. \\
%This section provides a detailed performance analysis of the different models built across the work.

%This section provides a detailed performance analysis of the different models built across the work.
\begin{table*}[!t]
\centering 
\scalebox{0.75}{
    \begin{tabular}{l c c c  c c c c c c c c c}
    \toprule
      {} & \multicolumn{3}{c}{\textbf{Overall}} &  \multicolumn{3}{c}{\textbf{Seen Categories}} & \multicolumn{3}{c}{\textbf{Unseen Entities}} & \multicolumn{3}{c}{\textbf{Unseen Categories}} \\
      \\
    {}  & F1 & Precision  & Recall & F1 & Precision  & Recall & F1 & Precision  & Recall & F1 & Precision  & Recall \\
    \midrule
    bt5 ($^{\diamondsuit \triangle}$) \cite{agarwal2020machine}   &  0.675 & 0.663 & 0.695 & 0.877 & 0.875 & 0.880 & 0.645 & 0.614 & 0.697 & 0.539 & 0.528 & 0.555 \\
    CycleGT ($^{\heartsuit}$) \cite{guo2020cyclegt}   & 0.309 & 0.306 & 0.315 & 0.545 & 0.538 & 0.558 & 0.179 & 0.178 & 0.182 & 0.181 & 0.179 & 0.183 \\
    \midrule
    T5-Baseline ($^{\diamondsuit \triangle}$) [our implementation] & 0.394 & 0.389 & 0.401 & 0.429 & 0.426 & 0.433 & 0.459 & 0.456 & 0.464 & 0.345 & 0.335 & 0.356 \\
    T5-Finetune ($^{\diamondsuit \triangle}$) [our implementation] & 0.179 & 0.179 & 0.182 & 0.171 & 0.170 & 0.173 & 0.175 & 0.175 & 0.178 & 0.201 & 0.201 & 0.205 \\
    \textbf{T5-MSSF} ($^{\diamondsuit \Join}$) [our framework] & 0.431 & 0.432 & 0.431 & 0.428 & 0.439 & 0.423 & 0.476 & 0.476 & 0.476 & 0.408 & 0.407 & 0.416 \\
    \bottomrule
    \end{tabular}
}
\caption{Multi-task (${\diamondsuit}$). Unspervised (${\heartsuit}$). Semi-Supervised ($\Join$). Supervised (${\triangle}$). Summary of related work results in Text-to-Graph (T2G) (in strict matching schema) with multi-task, supervised and/or unsupervised learning. Our multi-task semi-supervised framework is T5-MSSF.}
\label{tab:t2g-results}
\end{table*}

%TODO: pseudo-code 
\setlength{\parindent}{0pt}\textbf{Non-parallel Dataset}. Cycle training makes use of large unlabeled data, so we collected two datasets, one for natural English text and another one for graphs in English language. These datasets are extracted by means of our crawling-scraping algorithm which uses Wikipedia\footnote{\url{https://www.wikipedia.org}} and Wikidata\footnote{\url{https://www.wikidata.org}} for content retrieval as described:

\begin{enumerate}
    \item In the first iteration, we have to define an origin entity, but afterwards, the crawling algorithm will select the last entity (LIFO policy) following a Depth First Search strategy.
    \item Send query, SPARQL-based, to the Wikidata - API service to retrieve all the graphs in which their subject is the queried entity.
    \item Scrape at most the 4th first paragraphs of entity’s page in Wikipedia.
    \item For each object in the retrieved graph, go to the first step, unless it has reached a depth\footnote{This depth represents the number of parents already analysed w.r.t to the current entity.} of 5 or it has already been crawled.
\end{enumerate}

 By executing this crawling algorithm 4 times and pre-processing all data, we extracted a total of 176,000 unique entities with 532,288 triples and 240,024 paragraphs. Finally, we built two datasets, one for the graphs with 271,095 instances (1 to 6 triples per each), and another one for the natural text, with 240,024 instances (one sentence or more per each) of 459.67characters on average length.

\begin{comment}
    \begin{algorithm}
    \caption{Wikidata \& Wikipedia crawling following a Breadth-First Search policy for content extraction.}\label{alg:cap}
    \begin{algorithmic}[1]
    \Require{$y \gets List(Pair($Wikidata Id, Entity Name$))$}
    \Assume{$getGraph \ : \ List(Pair(E,V)) \gets$ Wikidata Id}
    \Assume{$getText \ : \ String \gets$ Entity Name}
    \While{$y$ is not empty}
    \State{$id, \ name \gets DEQUE(y)$}
    \State{$graph \gets getGraph(id)$}
    \State{$text \gets getText(name)$}
    \For{$edge, \ vertice$ in $graph$}
    
    \EndWhile
    \end{algorithmic}
    \end{algorithm}
\end{comment}

\setlength{\parindent}{0pt}\textbf{Models}. T5-Baseline is trained with all supervised samples from the WebNLG dataset. T5-MSSF is finetuned with 15\% of the supervised data, but then it is cycle-trained with 30,000 different synthetic samples on each iteration along the supervised ones. Similary, T5-Finetune is trained with 15\% of the supervised data, however, the cycle-training part is removed to asses the advantages of such approach.

\setlength{\parindent}{0pt}\textbf{Training Details}\label{exp:training}. We trained on mini-batches (8 samples) with accumulation steps (4 steps), at a learning rate of $2.0e^{-4}$ (fine-tuning) and $1.0e^{-5}$ (cycle training). The maximum number of epochs are set to 50 (fine-tuning) and 30 (cycle training) with 5 epochs of patience. The maximum source and target length is limited to 64 tokens with 4-beam search. Repetition and length penalty are applied, 2.5 and 1.0 respectively, along early stopping. The number of cycle steps are restricted to 3. For the sake of simplicity, each iteration of the cycle training is configured with the same hyper-parameters. 

\setlength{\parindent}{0pt}\textbf{Evaluation Metrics}. On one hand, G2T performance is studied with the following metrics: BLEU \cite{papineni2002bleu}, chrF++ \cite{popovic2015chrf}; the greater, the better performance, and TER\cite{snover2006study}; the lower, the better performance. On the other hand, T2G performance is analysed with the well-known metrics: F1, Precision and Recall, on strict measurement, i.e. it evaluates an exact match of the hypothetical triple with the reference triple.

\section{Results}\label{results}

As shown in Table \ref{tab:g2t-results}, for G2T generation in overall analysis, even with only 15\% of pairing information between text and graphs, our T5-MSSF model (multi-task semi-supervised) can achieve a 45.3 BLEU score, which improves the performance of our model trained on the full supervised data, T5-Baseline, and the CycleGT model (unsupervised). Similarly, T5-MSSF attains the best performance on  TER and chrF++ within our work, but fails to surpass performance from both CycleGT (unsupervised) and bt5 (multi-task) models on these metrics. There is an interesting pattern that reveals that unsupervised or semi-supervised learning perform more similar across the different data types (seen categories, unseen entities and unseen categories) than supervised ones. For instance, bt5 and T5-Baseline reach differences of 17.1 and 13.7 BLEU points respectively, between the best (seen categories) and the worst (unseen categories) performance of themselves. However, CycleGT and T5-MSSF obtain differences of 6.5 and 10.9 BLEU points respectively between the same data types. Probably, given the same performance on the overall, then, for unseen domains these unsupervised learning methods are more likely to generalize better. 

In Table \ref{tab:t2g-results}, T2G parsing, T5-MSSF reaches best performance among our models and surpasses again CycleGT in unsupervised learning by a 39.4\% F1-Score improvement in the overall analysis. In fact, T5-MSSF greatly exceeds (140.7\% F1-Score improvement) T5-Finetune performance, however, it is not yet capable of reaching bt5 performance. As previously observed, the performance of our multi-task semi-supervised model across the different data types is very similar, reaching at most a difference of 0.068 F1-Score points between the best (unseen entities) and the worst (unseen categories). Contrary, bt5 model presents a difference of 0.338 F1-Score points between the best (seen categories) and the worst (unseen categories), but even the unsupervised method, CycleGT, holds a difference of 0.364 F1-Score points between the same data types. This trend might suggests that synthetic data distribution in our non-parallel samples helps our framework to better generalize than using decoupled parallel data in which there exists a real distribution, as CycleGT does.

\section{Conclusion}\label{conclusions}

This paper proposed a multi-task semi-supervised framework based on the T5-Base model for Graph-to-Text \& Text-to-Graph. Our cycle framework improves unsupervised state-of-the-art resuts on both tasks using only 15\% of supervised examples, but in comparison to supervised models there is still room for improvement. This framework consists in translating text to graphs, and vice-versa, using the model itself, so it obtains parallel data from fully non-parallel samples. These samples are added (optionally) to the original parallel data before training the final model. This approach resulted to be very convenient to obtain similar performance across the different domains. Furthermore, the resulting model can be trained in any new domain with non-parallel data, by simply adding text and graphs in our framework. We release our collected non-parallel dataset. Thus, future work can research whether this framework would be useful for lifelong learning.

\flushend
\bibliographystyle{named}
\bibliography{ijcai22}

\end{document}